# Evaluation of the syllables pronunciation quality in speech rehabilitation through the solution of the classification problem

Evgeny Kostyuchenko[1][0000-0001-8000-2716]

[1] Tomsk State University of Control Systems and Radioelectronics, Tomsk 634050, Russia
key@fb.tusur.ru

**Abstract.** The solution of the problem of assessing the quality of the pronunciation of syllables during speech rehabilitation after surgical treatment of oncological diseases of the organs of the speech-forming tract is considered in the work. The assessment is carried out by solving the problem of classifying syllables into two classes: before and immediately after surgical treatment. A classifier is built on the basis of the LSTM neural network and trained on the records before the operation and immediately after it, before the start of speech rehabilitation. The measure of assessing the quality of syllables pronunciation in the process of rehabilitation is the metric of belonging to the class before the operation. A study is being made of the influence of taking into account problematic phonemes, the gender of the patient, his individual characteristics on the resulting estimates of the quality of pronunciation. A comparison with existing types of syllable pronunciation quality assessments is carried out, recommendations are given for the practical application of the resulting new class of pronunciation quality assessments.

## 1    Introduction

The problem of oncological diseases remains relevant for the last decades. This statement is true both for the whole world as a whole and for a single country, in particular, the Russian Federation. So, in the world in 2020, cancer was detected in 19.3 million people. 10 million people died as a result of this disease [1]. In Russia, similar figures are 556,036 new cases and 291,461 deaths, respectively. Despite the difficulties in obtaining accurate objective data in 2020 due to the onset of the pandemic and the difficulties of operational monitoring and preventive diagnostics, the increase in the incidence over the past decade is beyond doubt [2]. This fact is confirmed by the generalization of statistics for earlier years. The negative dynamics of indicators is shown in Figure 1.The general dynamics of changes in the incidence of these types of tumors per 100,000 over the past 10 years is presented in Figure 1, a. The dynamics of the total number of identified diseases is also negative. As can be seen from Figure 1, b, the total share of localization presented (Figure 1, c) and the cumulative risk of developing malignant neoplasms in the range of 0-74 ages (Figure 1, d) are growing.

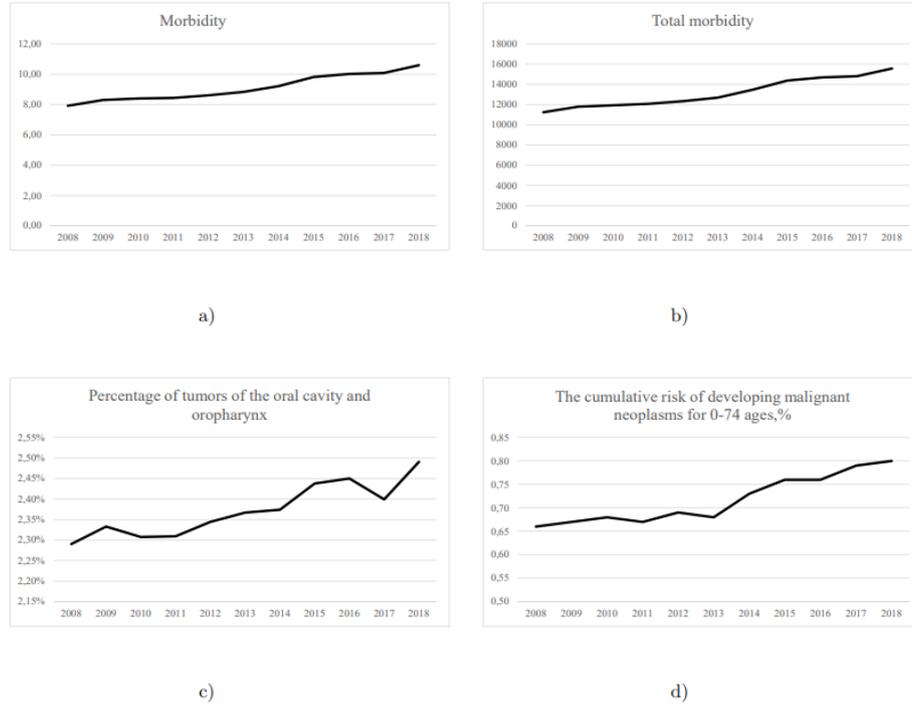

**Fig. 1.** The dynamics of the incidence of tumors of the oral cavity and oropharynx in 2008-2018 years [3, 4].

The given dynamics confirms the relevance of the problem of oncological diseases, in particular, with localizations in the region of the speech-forming tract. Surgical treatment leads to impaired speech function and the need for speech rehabilitation. At the same time, it should be noted that the speech function is one of the determining ones in assessing the quality of life, as it affects the patient's interaction with society and the possibility of working in many areas. When conducting speech rehabilitation, one of the key points is the assessment of the quality of pronunciation of speech units. At the same time, in a number of cases, it is the pronunciation that is important without taking into account the influence of the context on the recognition process. It is for this reason that the problem of assessing the quality on the minimum phonetic units is of interest. In this work, the syllable acts as such a unit. The work is devoted to the study of a new approach based on solving the problem of classifying syllables into two classes - before the operation and after the operation, and using the class membership metric as an assessment of the quality of pronunciation.

## 2 Related works

There are several different classes of approaches to the definition of speech quality. Subjective - based on the application of expert opinions in accordance with standards, for example, GOST 50840-95, MOS [5-6]. Direct subjective assessments require the participation of experts (for example, 5 people GOST - not feasible in practice) and have all the disadvantages of expert assessments. Estimates from the field of communication, for example, PESQ, E-Model and others [7-8]. As a rule, they lead to a comparison of the signal before and after transmission and are poorly adapted to the comparison of different implementations of signals with different durations and markings. Methods based on the use of different distances between signal realizations [9-10]. However, the use of distance-based metrics leaves questions about the interpretability and comparability of the resulting values, especially for different speakers. Separately, we can distinguish methods based on the use of speech recognition (in fact, the use of algorithms instead of experts within the framework of methods based on standards) [11-12]. However, the use of speech recognition requires the presence of significant phonetic units (preferably phrases), which leads to the influence of the context and the impossibility of excluding it when obtaining quality ratings.

Other works devoted to the assessment of speech quality can be somehow correlated with the given brief classification. They talk about the possibility of obtaining objective assessments in an automated or automatic mode. But the above shortcomings speak of the prospects of searching for new methods for assessing the quality of speech and also confirm the relevance of this work.

## 3 Statement of the problem and description of the approach

Previous studies [9, 13] have shown that the stage of the disease at the time of the patient's admission to a medical institution has practically no effect on the characteristics associated with the quality and intelligibility of speech. Based on this, a conclusion was made about the possibility of using speech recordings before surgical treatment as a standard for an individual patient. Let's designate this set as Set 1. Such selection of the standard allows us to ignore the patient's congenital speech defects. It becomes possible to focus on the return of not an abstract idealized speech function, but on bringing speech to the quality of speech of the same patient before surgery. This leads to the possibility of reducing the variability of the speech signals used, which leads to a simplification of their analysis and the focus of the speech rehabilitation procedure on a particular patient, which is inherent in the approach.

Immediately after undergoing surgical treatment and healing of the wounds associated with it, the patient proceeds to the procedure of speech rehabilitation. At this stage, a new recording of the patient's speech is performed (Set 2). This set of records is an illustration of the effect of surgery on speech quality. During the course of rehabilitation, as the training tasks are completed, the patient makes new sets of records (Set 3, 4, etc.). The goal of rehabilitation is to get these sets as close as possible to Set 1 and away from Set 2.

In this work, in contrast to the methods from Section 2, it is proposed to use the following idea. The first two sets before and immediately after the operation can be considered as two different classes - the ideal speech for this patient and its opposite. The rehabilitation process has the goal of translating speech from close to set 2 to close to set 1. For this reason, based on the first two sets, a classifier of syllable records can be built, and the resulting class membership metrics can be used as estimates of speech quality. In particular, then the formalized goal of rehabilitation is to ensure that the records belong to the class of Set 1, and not Set 2. Membership metrics can be used to form an assessment of the quality of the pronunciation of a set of syllables, in particular, if there is a normalization of membership values by the sum 1, such an estimate can simply be that the set being evaluated belongs to the class Set 1.

This approach is easy to understand and interpret the values obtained, in addition, it leaves room for research, since it does not impose restrictions on the choice of tools used when building a classifier.

## 4 Method for solving the classification problem and description of the input data

### 4.1 Method for solving the classification problem

There are many methods for solving classification problems. We can highlight the use of support vector machines [14], fuzzy logic [15], the k nearest neighbors method [16], and many others. Despite the widespread use of these methods, in the framework of solving classification problems related to speech processing, convolutional neural networks have proven themselves in the best way. They are successfully used to solve the problems of speech recognition [17], determining the speaker's personality [18], and recognizing human emotions [19].

Despite the presence of a large set of standard convolutional architectures that have proven themselves for use in a wide variety of classes of problems, this study uses a neural network based on LSTM [20]. This network is adapted to process signals of various durations and, with a simple structure with a small number of neurons, does not lead to retraining on limited data sets.

The neural network considered in the framework of the study has the following architecture:

```
model = Sequential()
model.add(LSTM(128, return_sequences=True, input_shape=X.shape[1:]))
model.add(LSTM(64))
model.add(Dense(64))
model.add(Activation('tanh'))
model.add(Dense(16))
model.add(Activation('sigmoid'))
model.add(Dense(1))
model.add(Activation('hard_sigmoid'))
```

Thus, in essence, the used neural network consists of two LSTM layers and three fully connected layers, successively reducing the number of outputs to one. In fact, this output ultimately contains an assessment of the quality of the syllables presented to the trained neural network. The Adam optimization algorithm is used as a learning algorithm, the Loss function is binary crossentropy.

### 4.2   Input data and preparation

The study uses a dataset collected during the treatment of 165 patients. In the process of preparation, users with less than two recording sessions were excluded from this general set (because it is impossible to build a classifying model from one record). After that, subsets of exactly 100 syllables selected from the GOST 50840-95 table and subsets of exactly 90 syllables based on the phonemes most susceptible to change were separated. As a result of these procedures, users with any sets deviating from the above were excluded (for example, with omissions of single syllables - a common mistake). 2 data sets were obtained: for 100 syllables - containing 22 recording sessions from 8 users, of which 7 men and one woman, a total of 2200 entries, and for 90 syllables - containing 85 recording sessions from 32 users, of which 21 men and 11 women, total 8500 records.

To build the model, only the first two sessions of each user are used. The remaining sessions are used for additional validation of the resulting classification model and its comparison with other evaluation methods (peer review).

Before being submitted for processing by the neural network, fragments that do not contain a signal exceeding the minimum threshold are filtered out from each record. Next, the Fourier transform is applied to obtain a spectrogram. Fragments of the same length are cut from the obtained spectrogram, the size of one fragment is 8x513, which are then transferred to the neural network.

The final size of the resulting data set based on the first pair of sessions is 102322x8x513 when all users are combined for 90 syllables.

Before training, 80% of the resulting set is separated into a training set, 20% into a test set.

## 5   Description of the experiment

Training was carried out on the complete data set received, on a set for patients of the same sex and for individual patients separately. Considering that the set is balanced (the number of records in the first and second sessions is exactly the same), then accuracy is used as the metric for training and assessing the quality of the resulting model, as the most simple and understandable for balanced sets.

The conducted training showed that none of the models is retrained, which makes it possible to skip the possible use of augmentation at this stage. The change in Loss-function and accuracy during the learning process for one of the patients is shown in Figure 2.

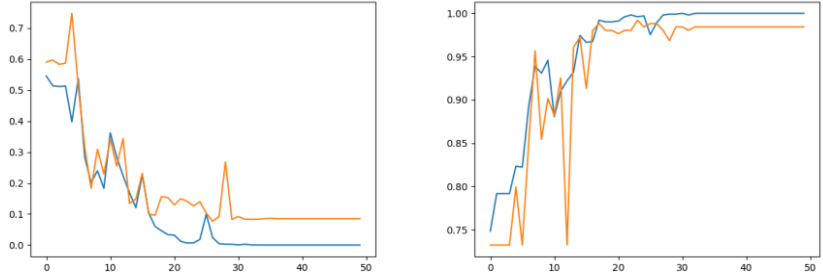

**Fig. 2.** Training and test for one patient. Loss function (left), accuracy (right)

Table 1 below shows the average accuracy values for various datasets.

**Table 1.** Values of the accuracy metric for various datasets.

|  | 90 syllables (problematic) | | 100 syllables (GOST) | |
| --- | --- | --- | --- | --- |
|  | Training | Test | Training | Test |
| Individual | 0.99 | 0.95 | 0.76 | 0.65 |
| Men | 0.60 | 0.55 | 0.57 | 0.52 |
| Women | 0.71 | 0.67 | 0.81 | 0.71 |
| All | 0.54 | 0.52 | 0.52 | 0.50 |

## 6   Discussion

Let's consider the received results of experiments.
1. 1. The neural network has difficulty classifying for users without separation. The resulting accuracy metric is slightly different from 0.5, which indicates the inoperability of the solution in this form.
2. Separation of users allows to slightly increase accuracy, however, the obtained metric values are still far from 1 even for the training sample, which does not allow using the obtained model for processing new signals - the results will have nothing to do with a real quality assessment pronunciation. Fundamental differences are visible only for the female voice for the selection of syllables from the GOST set. However, this jump is not caused by a significant increase in generalizing ability, but by the fact that only one patient is represented in this subsample. In essence, we move on to individual training of the model. The excess over the average value of an individual set is random.
3. The resulting higher values for women in the sample of 90 syllables are also explained by their smaller number.
4. The transition from the list of GOST syllables to the list of syllables based on problematic phonemes leads to an increase in accuracy. This is explained by the fact that the model has to work, firstly, with a smaller number of phonemes, and secondly, with those phonemes for which the differences between the classes are the strongest. Because of this, the division into classes is easier and more accurate.

Separately, we will compare the quality estimates obtained for one of the sessions in the process of undergoing rehabilitation, which was not used in the construction of the model. Let's compare the obtained values with the values obtained in the evaluation by experts according to GOST. The disadvantage of this approach for comparison is the fact that in the GOST methodology, estimates for individual signals are binary. In our case, they are continuous. Let's find the correlation coefficient for the estimates obtained by the expert and proposed methods. For one session of 90 records, it is 0.860, which indicates a strong agreement with the original method, taking into account identical ranges of quality scores (from 0 to 1 for each of the methods).

## 7     Conclusion

The obtained results allow us to conclude about the potential applicability of a new approach to assessing the quality of pronunciation of syllables based on solving the classification problem. This is supported by the high average value of the accuracy metric on the test sample. This is also confirmed by the correspondence of the obtained values and the absence of contradictions in comparison with the expert methodology. As a further study, it is necessary to compare with other methods for assessing the quality of pronunciation, in particular, based on the use of distances between syllable realizations. In addition, it is necessary to continue research in the field of application of the method not only to a single user, but also to sets of patients (in the general case, to the entire available data set). In addition, the results were obtained on a separate neural network architecture. In the future, it is planned to choose from a variety of architectures and justify the number of neurons in the intermediate layers based on the results of experiments.

## 8     Acknoledgements

This research was funded by the Ministry of Science and Higher Education of the Russian Federation within the framework of scientific projects carried out by teams of research laboratories of educational institutions of higher education subordinate to the Ministry of Science and Higher Education of the Russian Federation, project number FEWM-2020-0042.

The authors would like to thank the Irkutsk Supercomputer Center of SB RAS for providing access to the HPC-cluster «Akademik V.M. Matrosov» [21].